\pdfoutput=1

\documentclass[11pt]{article}

\usepackage{acl}

\usepackage{times}
\usepackage{latexsym}

\usepackage[T1]{fontenc}

\usepackage[utf8]{inputenc}

\usepackage{microtype}

\usepackage{algorithm}
\usepackage{algorithmic}
\usepackage{booktabs}
\usepackage{graphicx}
\usepackage{amssymb}
\usepackage{bbold}
\usepackage{bbm}
\usepackage{amsmath}
\usepackage{url}
\usepackage[switch]{lineno}  %
\usepackage{newfloat}
\usepackage{float}
\usepackage{subfig}

\title{Label Anchored Contrastive Learning for Language Understanding}

\author{Zhenyu Zhang, Yuming Zhao, Meng Chen, Xiaodong He\\
        JD AI, Beijing, China \\
        \texttt{\{zhangzhenyu47,zhaoyuming3,chenmeng20,xiaodong.he\}@jd.com}\\
        }

\begin{document}
\maketitle
\begin{abstract}
Contrastive learning (CL) has achieved astonishing progress in computer vision, speech, and natural language processing fields recently with self-supervised learning. However, CL approach to the supervised setting is not fully explored, especially for the natural language understanding classification task. Intuitively, the class label itself has the intrinsic ability to perform hard positive/negative mining, which is crucial for CL. Motivated by this, we propose a novel label anchored contrastive learning approach (denoted as LaCon) for language understanding. Specifically, three contrastive objectives are devised, including a multi-head instance-centered contrastive loss (ICL), a label-centered contrastive loss (LCL), and a label embedding regularizer (LER). Our approach does not require any specialized network architecture or any extra data augmentation, thus it can be easily plugged into existing powerful pre-trained language models. Compared to the state-of-the-art baselines, LaCon obtains up to 4.1\% improvement on the popular datasets of GLUE and CLUE benchmarks. Besides, LaCon also demonstrates significant advantages under the few-shot and data imbalance settings, which obtains up to 9.4\% improvement on the FewGLUE and FewCLUE benchmarking tasks.
\end{abstract}

\section{Introduction}

In recent years, contrastive learning (CL) has been widely applied to self-supervised representation learning and led to major advances across computer vision (CV) \cite{Moco2019, chen2020simple}, speech \cite{contrast-audio-general-repr/icassp/SaeedGZ21,seq-cpq-asr/icassp/ChenZWFKWWMWX21}, and natural language processing (NLP) \cite{cert21, simcse21, ConSERT21}. The basic idea behind these works is to pull together an anchor and a ``positive'' sample in the embedding space, and to push apart the anchor from many ``negative'' samples. Since no labels are available, a positive pair often consists of data augmentations of the sample (a.k.a ``views''), and negative pairs are formed by the anchor and randomly chosen samples from the mini-batch. In visual representations, an effective solution to generate data augmentations is to take two random transformations of the same image (e.g., cropping, flipping, distortion and rotation) \cite{chen2020simple,byol20,simclrchen2020big}. For natural language, similar approaches are adopted such as word deletion, reordering, substitution, and back-translation etc. \cite{cert21, CLINE21} However, data augmentation in NLP is inherently difficult because of its discrete nature. Therefore, some previous works \cite{simcse21, ConSERT21} also use dropout technique \cite{dropout14} to obtain sentence augmentations. 

Unlike self-supervised setting, some researchers propose supervised contrastive learning (SCL) \cite{scl20, sclce21,lcl21} which can construct positive pairs by leveraging label information. Examples from the same class are pulled closer than the examples from different classes, leveraging the semantics of labels to construct negatives and positives rather than shallow lexical information via data augmentation.
Despite the aforementioned advantages brought by SCL, we argue that CL under supervised learning is not fully explored because the label information can be better utilized. On the one hand, labels are usually not merely categorical indices in the label vocabulary, but also contain specific semantic meanings, especially in the language understanding tasks. 
Thus labels can be used as positive/negative samples or anchors when calculating contrastive loss. On the other hand, label embedding enjoys a built-in ability to leverage alternative sources of information related to labels, such as class hierarchies or textual descriptions. Once we obtain representative label embeddings, they can be utilized to enhance the image/text representations, and finally facilitate the classification task.  Previous label embedding based classification models \cite{wang2018joint,lsan-xiao2019label} have demonstrated the effectiveness of leveraging label information. 

Motivated by above analysis, we propose a novel label anchored supervised contrastive learning approach (denoted as LaCon), which combines the advantages of both contrastive learning and label embedding techniques. Specifically, we have the following three novel designs: 1) Instance-centered contrastive loss (ICL), which uses the InfoNCE \cite{infoNCE18} to encourage each text representation and its corresponding label representation to be closer while pushing far away mismatched instance-label pairs. We further apply a multi-head mechanism to catch different aspects of text semantics. 2) Label-centered contrastive loss (LCL), which takes label as anchor, and encourages the label representation to be more similar to the corresponding instances belonging to the same class in a mini-batch than the instances with different labels. 
3) Label embedding regularizer (LER), which keeps the inter-label similarity as low as possible thus the feature space of each class is more dispersed to prevent representation degeneration. By combining above three losses, LaCon can learn good semantic representations within the same space for both input instances and labels. It's also well aligned with the two key properties related to CL: alignment and uniformity \cite{ctl-theory20}, where alignment favors encoders that assign similar features to similar samples. Uniformity prefers a feature distribution that preserves maximal information, i.e., the uniform distribution on the unit hypersphere. 

To validate the effectiveness of LaCon, we perform extensive experiments on eight language understanding tasks.
We take the popular pre-trained language model BERT-base \cite{devlin2018bert} as text encoder without loss of generality. For simplicity, we predict the classification label by matching the instance representation with label embeddings directly. Since our approach does not require any specialized network architecture or any extra data augmentation, LaCon can be easily plugged into other pre-trained language models.
Additionally, we also explore the capability of LaCon under more difficult task settings, including few-shot learning and data imbalance situations. 

To summarize, our contributions are as follows:

\begin{itemize}
\item We propose a novel label anchored contrastive learning approach for language understanding, which is equipped with a multi-head instance-centered contrastive loss, a label-centered contrastive loss, and a label embedding regularizer. All three contrastive objectives help the model learn the joint semantic representations for both input instances and labels.

\item We conduct extensive experiments on eight public language understanding tasks from GLUE \cite{2019-GLUE} and CLUE \cite{2020-Clue} benchmarks, and experimental results show the competitiveness of LaCon. Additionally, we also experiment on more difficult settings including few-shot learning and data imbalance situations. LaCon experimentally obtains up to 9.4\% improvement over BERT-base on FewGLUE \cite{fewglue} and FewCLUE \cite{fewclue} benchmark tasks.

\item We analyze the contribution of each ingredient of LaCon, and also visualize the learned instance and label representations, showing the necessity of each loss component and the advantage of LaCon on representation learning over BERT fine-tuned with cross entropy.

\end{itemize}

\section{Model}

\begin{figure*}[!htb]
\begin{center}
\includegraphics[scale=0.25]{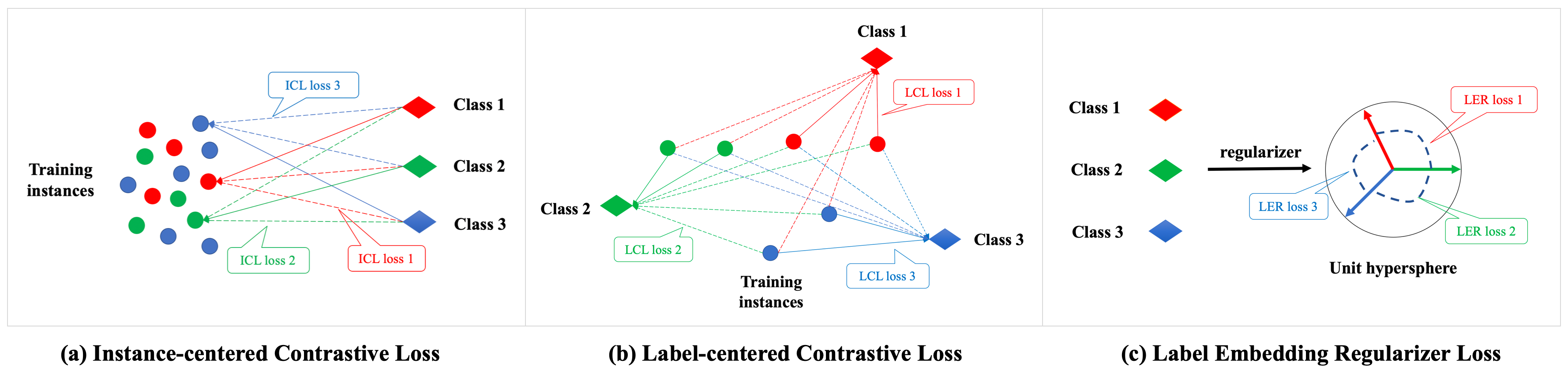}
\caption{Overview of LaCon. The full line is the similarity between a instance and corresponding label, and the dash line is the similarity between the mismatched instance and label. The lines with the same color denote the per-instance or per-label loss.}
\label{fig:LaCon}
\end{center}
\vspace{-3mm}
\end{figure*}

In this section, we introduce the details of LaCon. We focus on the language understanding classification tasks. For a multi-class classification problem with $C$ classes, we work with a batch of training examples $\{x_i, y_i\}$, where $1 \le i \le N$ and $1 \le y_{i} \le C$. Our target is to learn discriminative representations for both instances and class labels. As Figure \ref{fig:LaCon} shows, we propose three supervised CL based objectives, including the instance-centered contrastive loss, the label-centered contrastive loss, and the label embedding regularizer loss.

\subsection{The Input Encoder}
\label{sec:input-repr}

The input of LaCon contains two parts consisting of the text and all the labels for the task. Since SOTA language understanding classification models follow the ``pre-training then fine-tuning'' two-stage paradigm, here we take the prevalent pre-trained language model (PtLM) as input encoder. In this paper, we select BERT-base \cite{devlin2018bert} as the backbone for PtLM (denoted as $f$) without loss of generality. Given a text $\mathbf{x}=\{w_1,w_2,...,w_M\}$ containing $M$ tokens, the output of PtLM (i.e. BERT) is $\mathbf{E} = f_{PtLM}([CLS], w_1,..., w_M)$ where the $[CLS]$ token is the inserted sentence representation token and $\mathbf{E} \in R^{(M+1) \times d}$. $d$ is the dimension of the model. We use the first token output $E_{[CLS]}\in R^{d}$ to represent the whole input text. We then apply a projection head (denoted as $g$) that is a 3-layer MLP with ReLU activation function for each hidden layer to the $E_{[CLS]}$, where the dimension of the $g$ is also $d$. For a mini-batch $X$ with $N$ training samples, the text representations can be obtained as Equation \ref{eq:text-repr}.
\begin{equation}
\label{eq:text-repr}
\begin{split}
    H= g \circ f_{PtLM} (X), \,\, \textrm{where} \\
    X={x_{1}, x_{2},..., x_{N}} \,\, and \,\, H \in R^{N \times d}
\end{split}
\end{equation}

Along with each mini-batch, LaCon also maps the $C$ classes into $C$ label embeddings. For simplicity, we look up in a learnable weight matrix $W_{emb}$ and map the $k^{th}$ label into the $k^{th}$ row of $W_{emb}$, where the $W_{emb}$ is randomly initialized. The dimension $d$ is the same as PtLM. We then normalize the vectors for both $L$ and $H$ using the $l_2$ norm.
\begin{equation}
\label{eq:label-repr}
\begin{split}
    L= lookup([1, 2, ..., C], W_{emb}) \\ \textrm{where}\,\, L \in R^{C \times d}\,\, and \,\, W_{emb} \in R^{C \times d}
\end{split}
\end{equation}

\subsection{Instance-centered Contrastive Loss}
\label{sec:icl_loss}

Given a mini-batch of input text and corresponding label $(x_i, y_i)$, as shown in Figure \ref{fig:LaCon} (a), the instance-centered contrastive loss (ICL) takes each instance $x_i$ as anchor, and mines positive and negative samples from class labels. ICL aims to encourage each text representation and corresponding label representation to be closer while pushing far away mismatched instance-label pairs. As shown in Equation \ref{eq:inst-loss}, we modify the InfoNCE \cite{infoNCE18} to calculate the loss. Here we leverage the cosine similarity function as the distance metric $sim$. $\tau$ is the temperature hyper-parameter, which can be tuned to improve the performance. Lower temperature increases the influence of examples that are harder to separate, effectively creating harder negatives. Similar to the self-supervised CL, ICL also takes only one positive and many negatives. Differently, the positive is not generated from data augmentation, and the negatives are not randomly sampled from the same mini-batch. By treating the class labels as data samples, ICL can mine better positive and negatives with the supervision signal. By minimizing the ICL, the instance representation is aligned to its label representation in the same semantic space, which encourages the model to learn a more representative embedding for each class label. 

\begin{small}
\begin{equation}
\label{eq:inst-loss}
\resizebox{0.75\linewidth}{!}{
\begin{minipage}{\linewidth}
$
\begin{aligned}
    \mathcal{L}_{ICL} = -\frac{1}{N}\sum_{x_i,y_i}log \frac
    {exp(sim(H_{x_i},L_{y_i} )/\tau)}
    {\sum_{1 \leq p \leq C}
    exp(sim(H_{x_i},L_{p})/\tau )}
\end{aligned}
$
\end{minipage}}
\end{equation}
\end{small}

Inspired by the image-augmented views proposed in CV \cite{Moco2019, chen2020simple, simclrchen2020big, byol20}, 
we also leverage the multi-head mechanism proposed in Transformer \cite{transformer} to compute the ICL for each head representation with smaller representation dimension. Each head can be regarded as a clipped local view of the instance or label representation.  Suppose we have $m$ heads for both instance representation and label representation, then for the $k_{th}$ head, the corresponding representations for training sample ($x_i$, $y_i$) are $h_{x_i}^{k}$ and $l_{y_i}^{k}$, and the dimension of each vector becomes $d'= d/m$. Then, we apply the contrastive loss for each head by following Equation \ref{eq:mh-loss}. Compared with InfoNCE, $\mathcal{L}_{ICL}$ and $\mathcal{L}_{ICL}^{'}$ do not suffer from small batch size issue \cite{Moco2019,simclrchen2020big} because we only need to contrast the instance representation with corresponding label representation for per example loss. 

\begin{small}
\begin{equation}
\label{eq:mh-loss}
\resizebox{0.75\linewidth}{!}{
\begin{minipage}{\linewidth}
$
\begin{aligned}
\mathcal{L}_{ICL}^{'}= -\frac{1}{N}\sum_{k=1}^{m} \sum_{x_i,y_i}log \frac{exp(sim(h_{x_i}^{k},l_{y_i}^{k} )/ \tau )}{\sum_{1 \leq p \leq C}exp(sim(h_{x_i}^{k},l_{p}^{k})/\tau )}
\end{aligned}
$
\end{minipage}}
\end{equation}
\end{small}

\subsection{Label-centered Contrastive Loss}
\label{sec:lcl_loss}

As shown in Figure \ref{fig:LaCon} (b), we can take the class label in a mini-batch as anchor, and mine positive/negative samples from corresponding instances. Suppose there are $|P|$ classes in the batch, where $P=\{p| 1 \le p \le C \wedge |A(p)|>0 \}$. We define that $A(p)$ denotes the set of indices of all positive instances whose label is $p$, i.e. $A(p)=\{x_i| y_i = p\}$. And $B(p)$ represents the set of negative instances whose label is not $p$, i.e. $B(p)=\{x_j| y_j \neq p \}$. Then we can calculate the label-centered contrastive loss (LCL) as Equation \ref{eq:label-loss}, which promotes the instances of a specific label to be more similar than the others for each label. Similar to the previous SCL \cite{scl20, sclce21}, LCL also contains many positives per anchor and many negatives. Different from SCL which sums up all the softmax scores among all pairs of instances of the same class in a batch, LCL is based on comparing a specific label representation with corresponding instances (i.e. $A(p)$). LCL is more stable as the label representation serves as the anchor which can be stably updated.

\begin{small}
\begin{equation}
\label{eq:label-loss}
\resizebox{0.75\linewidth}{!}{
\begin{minipage}{\linewidth}
$
\begin{aligned}
\mathcal{L}_{LCL} = -\frac{1}{|P|}\sum_{p \in P} \sum_{a \in A(p)} log \frac{exp(sim(L_{p}, H_{a})/ \tau )}{\sum_{b \in B(p)} exp(sim(L_{p}, H_{b})/\tau )}\
\end{aligned}
$
\end{minipage}}
\end{equation}
\end{small}

ICL and LCL are complementary to each other and more computationally efficient than previous SCL. We conduct the detailed theoretical analysis in Appendix \ref{app:theory} due to space limitation.

\subsection{Label Embedding Regularizer}
\label{sec:ler_loss}
Recent researches \cite{ctl-theory20} demonstrate that it is common and useful to add a regularization term during training to eliminate the anisotropy problem. Inspired by this, We devise a label embedding regularizer as shown in Equation \ref{eq:loss-reg} to promote the uniformity of our model and prevent model degeneration.
As illustrated in Figure \ref{fig:LaCon} (c), the label embedding regularizer (LER) encourages the label representations to be dispersed in the unit hypersphere uniformly. The LER loss is the exponential mean of the cosine similarity for all pairs of label representations. As $-1 \le sim(L_i, L_j) \le 1$, it is quite sensitive to the loss change as the gradient is larger than 1 for $exp(x)$ w.r.t $x \ge 0$. Thus, we add 1.0 to the cosine similarity so that the value of LER varies from 0 to $e^2$ - 1. 
\begin{equation}
\label{eq:loss-reg}
\resizebox{0.75\linewidth}{!}{
\begin{minipage}{\linewidth}
$
\begin{aligned}
\mathcal{L}_{LER} = avg(\sum_{i \ne j}(exp(1.0+sim(L_{i},L_{j}))-1.0) 
\end{aligned}
$
\end{minipage}}
\end{equation}

Finally, the overall loss function of LaCon is summarized as follows:
\begin{equation}
\label{eq:loss-LaCon}
\begin{split}
\mathcal{L} = \mathcal{L}_{ICL}^{'} + \mathcal{L}_{LCL} + \lambda * \mathcal{L}_{LER}
\end{split}
\end{equation}
where $\lambda$ is a hyper-parameter to balance the influence of our regularization term.

\subsection{Matching based Class Prediction}
\label{sec:matching_layer}
Since LaCon is capable of learning instance and label representations jointly, we can predict the class by matching the instance representation to all label representations directly during inference, just as shown in Equation \ref{eq:pred}. We denote this simple and direct approach as \textbf{LaCon-vanilla}. $H_{x}$ is the instance representation and $L_{j}$ is the label representation of Class $j$. $sim$ denotes the cosine similarity and $1 \le j \le C$ denotes the corresponding label. The advantage of LaCon-vanilla is that it does not require any complicated network architecture and can be easily plugged into the mainstream PtLMs. As a result, our inference-time model contains exactly the same number of parameters as the model using the same encoder but trained with cross entropy loss. 
\begin{equation}
\label{eq:pred}
pred=\arg_{max}(j)\{score_{j}|sim(H_{x},L_{j})\}
\end{equation}

\subsection{LaCon with Label Fusion}
\label{sec:explicit-fuse}
Previous researches \cite{Akata2016, wang2018joint, lsan-xiao2019label, Pappas2019, Miyazaki2020} have proved that incorporating the label semantics into the sentence representation can improve the model performance because the label information can highlight the alignment of input tokens and label information via carefully designed fusion mechanism. Inspired by LEAM \cite{wang2018joint}, here we design a fusion block to enhance the instance representations by utilizing the learnt discriminative label embeddings. We firstly calculate the cosine similarity interaction matrix $G$ between words and labels, and then apply a convolution then max-pooling layer ($conv_{max}$) to measure the attention score ($\beta_i$) for each word attending the instance representation. The fusion process is illustrated as Equation \ref{eq:fusion-leam}. Then the fused vector $z$ is fed into the projection head $g$ to get the enhanced instance representation.

\begin{small}
\begin{equation}
\label{eq:fusion-leam}
\begin{split}
m=conv_{max}(G),\, \textrm{where}\,\, G_{ij}=\frac{ <L_i, E_j> }{ ||L_{i}||\cdot||E_{j}|| }
\\
z=\sum_{i}\beta_iE_i,\, \textrm{where} \, \, \beta =softmax(m)\\
\end{split}
\end{equation}
\end{small}


To distinguish with the vanilla model above, we name this approach as \textbf{LaCon-fusion}. Please note that the fusion block is just applied between the text encoder and projection head, so the class prediction keeps the same as the LaCon-vanilla. Since the fusion block is not the main focus of this paper, we leave exploring more advanced fusion networks to future work.

\begin{table}[htp]
\centering
\small
\begin{tabular}{cccccc}
\hline \textbf{Datasets} & \textbf{type} & \textbf{class} & \textbf{metric} & \textbf{train} & \textbf{dev}\\ \hline

  {DBPedia}  & {genre}  & {14} & {ACC} & {14K} & {70K} \cr
  {Tnews}  & {genre}  & {15} & {F1} & {14.2K} & {10K} \cr
  {QQP}  & {PI}  & {2} & {ACC} & {10K} & {40.4K} \cr
 {MRPC}  & {PI}  & {2} & {ACC} & {4.07K} & {1.7K} \cr
 {QNLI}  & {NLI}  & {2} & {ACC} & {10K} & {5.4K} \cr
 {RTE}  & {NLI}  & {2} & {ACC} & {2.5K} & {278} \cr
 {CoLA}  & {LA}  & {2} & {M' corr} & {8.5K} & {1K} \cr
 {YelpRev}  & {senti}  & {2} & {ACC} & {10K} & {10K} \cr

\hline
\end{tabular}
\caption{The statistics of datasets that are from GLUE \cite{2019-GLUE} and CLUE \cite{2020-Clue}}
\label{tab:datasets}
\end{table}

\section{Experiments}
\begin{table*}[htp]
\centering
\small
\begin{tabular}{c|cccccccc}
\hline \textbf{Methods} & \textbf{YelpRev}  & \textbf{DBPedia}  & \textbf{Tnews} &  \textbf{QNLI} & \textbf{RTE} & \textbf{QQP} & \textbf{MRPC} & \textbf{CoLA}  \\ \hline

CE   & {82.0$\pm 0.5$} & { 98.7$\pm0.3$ } & {54.5$\pm 0.3$}  & {87.1$\pm 0.2 $} & {67.3$\pm 1.9 $} &  {82.2$\pm 0.5 $} & {85.6$\pm 1.6 $} & {60.9$\pm 0.8 $} \cr

LEAM  & {82.1$\pm 0.6 $} & {98.7$\pm0.5$} & {54.1$\pm 0.3 $}  & {87.2$\pm 0.7 $} & {67.3$\pm 1.3 $} & {81.9$\pm 0.5 $} & {85.6$\pm 1.3 $} &  {60.9$\pm 1.0 $} \cr

LSAN  & {82.2$\pm 0.6 $} & {98.7$\pm0.7$} & {54.9$\pm 0.8  $} & {87.1$\pm 0.3 $} & {69.7$\pm 1.0 $} & {81.2$\pm 0.5 $} & {86.1$\pm 0.7 $} & {61.6$\pm 0.9  $} \cr

CE+CL & {82.2$\pm0.6$}  & {98.5$\pm0.5$} & {53.9$\pm0.5$} & {87.3$\pm0.3$} & {67.8$\pm1.5$} & {82.4$\pm0.3$} & {83.1$\pm0.7$} & {61.1$\pm0.7$} \cr

CE+SCL  & {81.4$\pm 0.8 $}  & {98.5$\pm0.6$} & {54.6$\pm 0.2 $}  &  {87.7$\pm 0.1 $} & {69.1$\pm 2.2 $}
& {82.5$\pm 0.6 $} & {88.1$\pm 0.9 $} & {62.3$\pm 0.6 $} \cr

\hline
LaCon-vanilla  &  {82.3$\pm 0.5 $}  & {98.9$\pm0.5$} & \textbf{56.8$\pm \textbf{0.6} $}  & {88.1$\pm 0.2 $} & {71.4$\pm 0.7 $} & {82.8$\pm 0.5 $} & {87.5$\pm 0.9 $} & {62.4$\pm 1.0 $} \cr

LaCon-fusion &  \textbf{83.1$\pm \textbf{0.8} $} & \textbf{99.5$\pm \textbf{0.2}$} & {56.7$\pm 0.3$}  & \textbf{88.4$\pm \textbf{0.3}$} & \textbf{72.2$\pm \textbf{0.9}$} & \textbf{83.7$\pm \textbf{0.5}$} & \textbf{88.6$\pm \textbf{0.7}$} & \textbf{62.8$\pm\textbf{0.5}$} \cr

\hline
\end{tabular}
\caption{The experimental results for the Language Understanding Tasks. 
Best scores for each dataset are highlight in \textbf{bold} (all with significance value $p < 0.05$).
}

\label{tab:exp-all}
\vspace{-3mm}
\end{table*}

\subsection{Experimental Setup}
\label{sec:exp-setup}

\textbf{Datasets.} We experiment on 8 public datasets listed in Table \ref{tab:datasets}. which are from GLUE \cite{2019-GLUE}, CLUE \cite{2020-Clue}, DBpedia, and Yelp Dataset Challenge 2015. They cover five representative tasks including sentiment analysis (senti), genre classification (genre), paragraph identification (PI), natural language inference (NLI) and linguistic acceptability (LA). To improve the comparability and experiment confidence of the models, we follow the experimental setup in \cite{mixtext-txtcls/acl/ChenYY20} and use part of the training sets via sampling and the full original test sets for evaluation. We randomly sample without replacement at most 5K (binary-class) / 1K (multi-class) training instances per class from the whole datasets except for MRPC, RTE, and CoLA. 
We use the wilcoxon rank test \cite{wilcoxon1945individual} to check the statistic significance. The results of 10 runs are reported for each dataset in the format as ``avg$\pm std.dev$". 

\noindent \textbf{Training \& Evaluation.} During training, we run experiments for MRPC, RTE and CoLA with 10 random seeds on the whole training datasets and run the sampling strategy with 10 repeats for the remaining datasets. The average evaluation metrics are reported to avoid the noise and unstable randomness of a single run. 
We use the AdamW optimizer with initial learning rate as \{1e-5, 2e-5, 3e-5\} with linear learning scheduler, 6\% of warm-up steps of total optimization steps, and batch size as \{8,16,32,64,96\}, where the hyper-parameters are tuned for different datasets. For evaluation, we leverage accuracy (ACC), Macro-F1 score and Matthew's corr (M' corr) metrics to evaluate the performance. We run 10 epochs for all the datasets\footnote{We split 5\% of training set as validation for early stop.} and then evaluate the models on dev set. Our implementation is based on Huggingface Transformers\footnote{\url{https://github.com/huggingface/transformers}}. 

\subsection{Baselines}
Since LaCon is based on CL and label embedding technique, we compare with several SOTA models in language understanding including BERT-base fine-tuned with cross-entropy (CE) loss, label embedding based models, and self-supervised CL and supervised CL models.
\begin{itemize}
    \item \textbf{CE}: we directly follow the instructions of original paper \cite{devlin2018bert} to finetune BERT for both English and Chinese langauge understanding tasks.
    
    \item \textbf{LEAM}: \citet{wang2018joint} apply cosine similarity to get matching scores between words and labels and use CNN on the matching matrix to get the label-aware attention weighted text representation for classification.
    
    \item \textbf{LSAN}: \citet{lsan-xiao2019label} propose a label specific attention network that leverages label-attention and self-attention mechanism with an adaptive attention fusion strategy for multi-label classification. We use \textit{softmax} instead of \textit{sigmoid} for the model output due to our multi-class classification setting.
    
    \item \textbf{CE+CL}: \citet{ConSERT21} propose to learn sentence representations by joint fine-tuning PtLM with InfoNCE and cross-entropy based on feature augmentation. Here we leverage the framework of ConSERT \cite{ConSERT21} and the feature augmentation in SimCSE \cite{simcse21} to finetune and evaluate the PtLM on classification datasets.
    
    \item \textbf{CE+SCL}: \citet{sclce21} propose to boost sentence representation learning by fine-tuning PtLM with both supervised contrastive learning loss \cite{scl20} and cross entropy loss. We follow the instructions in \cite{sclce21} to set hyper-parameters.
    
\end{itemize}
\textbf{LEAM} and \textbf{LSAN} are label embedding based methods while \textbf{CE+CL} and \textbf{CE+SCL} are contrastive learning based methods. To compare all models fairly, we use BERT-base encoder for all the baselines and our proposed model.

\subsection{Main Results}
We report the experimental results of eight language understanding tasks in Table \ref{tab:exp-all}. It's observed that, LaCon-vanilla outperforms all the baselines in 7 datasets except MRPC, and LaCon-fusion achieves the best performance across all datasets. Specifically, 1) LaCon-vanilla outperforms BERT fine-tuned with CE by 4.1\%, 2.3\%, 1.9\%, and 1.5\% on RTE, Tnews, MRPC, and CoLA respectively, which indicates our proposed novel CL approach can facilitate the representation learning; 2) Compared with previous supervised contrastive learning method (CE+SCL), LaCon-fusion can still obtain very exciting improvements of 3.1\%, 2.1\%, 1.7\%, 1.2\% points on RTE, Tnews, YelpRev, QQP, which demonstrates the label fusion block can enhance the instance representations effectively; 3) Compared to previous label embedding methods (LEAM and LSAN) which are also equipped with the label fusion block, LaCon-fusion outperforms them with a large margin, which proves that LaCon can learn more discriminative joint representations for both labels and instances.



\subsection{Ablation Study}
\label{sec:ablation}

In this section, we conduct three groups of ablation studies to investigate the contribution of each component in LaCon. We only conduct experiments on MRPC, RTE, and CoLA datasets due to space limitation. The experimental results are shown in Table \ref{tab:exp-gradient}. First, we replace the multi-head ICL with single head version (LaCon w/ $\mathcal{L}_{ICL}$). Table \ref{tab:exp-gradient} shows the performance drops on all three datasets. We conjecture that the multi-head version can learn different parts of the local features of the representation, which can catch the text semantics in more fine-grained granularity. Second, we remove each of our proposed CL loss separately, and the results in the second part of Table \ref{tab:exp-gradient} demonstrate that ICL plays a more important role while LCL and LER are complementary to further improve the performance. We also try to add each CL loss in an accumulative way, please refer to Appendix \ref{app:accum_ablation} for more details.
Finally, we try to remove the projection head $g$ from LaCon and the performance degrades significantly, which indicates $g$ is critical in CL. Previous researches \cite{simclrchen2020big} also find the projector head can eliminate the non-task relevant features of the encoder in CL and benefit the downstream tasks. Meanwhile, Table \ref{tab:exp-gradient} shows that it is basically useless by adding $g$ to BERT directly (BERT w/ $g$), indicating that the projector head needs to be used with CL.  

\begin{table}[htp]
\centering
\small
\begin{tabular}{ccccc}

\hline \textbf{Methods} & \textbf{MRPC} & \textbf{RTE} & \textbf{CoLA} \\ \hline
LaCon-vanilla &  \textbf{87.5$\pm \textbf{0.8} $} & \textbf{71.4$\pm \textbf{0.7} $} & \textbf{62.4$\pm \textbf{1.1} $}  \cr
LaCon w/ $\mathcal{L}_{ICL}$ & {87.0$\pm 1.2 $} & {69.2$\pm 1.4$} & {61.5$\pm 0.9 $}  \cr

\hline
$-\mathcal{L}_{ICL}^{'}$ & {86.6$\pm 1.3 $} & {68.1$\pm 0.9 $} & {61.3$\pm 0.7 $}  \cr
$-\mathcal{L}_{LCL}$ & {87.1$\pm 0.6 $} & {70.5$\pm 0.8$} & {61.2$\pm 1.1 $} \cr
$-\mathcal{L}_{LER}$ & {87.3$\pm 1.1 $} & {70.2$\pm 1.3$} & {62.2$\pm 1.6$}  \cr

\hline
$-g$ & {86.8$\pm 0.6 $} & {69.6$\pm 0.6$} & {62.2$\pm 0.9 $} \cr
BERT w/ $g$ & {84.9$\pm 1.7 $} & {66.5$\pm 2.1$} & {61.0$\pm 1.2 $} \cr
\hline

\end{tabular}
\caption{Ablation study. Best scores for each dataset are highlight in \textbf{bold} (all with significance test $p < 0.05$).}
\label{tab:exp-gradient}
\end{table}
\section{Discussion}
In this section, we conduct further experiments under more challenging few-shot and data imbalance settings. We also discuss the hyper-parameter tuning and the impact of class number on LaCon. 

\subsection{LaCon for Few-shot Learning}

Few-shot learning is critical for applications of language understanding models because the high-quality human annotated datasets are usually costly and limited. Previous researches \cite{rdrop21, sclce21, betterft21} find that fine-tuning PtLM with cross entropy loss in NLP tends to be unstable across different runs especially when supervised data is limited. This limitation can result in model degeneration and model shift. Besides, some researches \cite{lbs19} also demonstrate that the cross-entropy optimization goal is not reachable due to the bounding of the gradient, which can also easily result in overfitting. Since LaCon is equipped by CL, it's interesting to validate if LaCon can overcome the shortcomings of CE under few-shot learning settings.

\begin{table}[htp]
\centering
\small
\begin{tabular}{ccccc}
\hline
\textbf{Model} & \textbf{YelpRev} & \textbf{Tnews} & \textbf{EPRSTMT} & \textbf{BUSTM} \cr
 \hline
CE & {59.0} & {52.5} & {84.4} & {65.6}  \cr
LaCon & \textbf{65.0}  & \textbf{55.8} & \textbf{90.6} & \textbf{75.0} \cr
 \hline
\textbf{Model} & \textbf{QNLI} & \textbf{RTE} & \textbf{MRPC} & \textbf{QQP}   \\ 
 \hline
CE  & {73.0} & {54.0} & {65.8} & {64.0}   \cr
LaCon & \textbf{76.0} & \textbf{60.0} & \textbf{69.0} & \textbf{71.0}  \cr
\hline
\end{tabular}
\caption{Performance under few-shot learning settings.}
\label{tab:exp-fewshot}
\end{table}

We conduct further experiments with vanilla LaCon on 5 public English datasets from FewGLUE \cite{fewglue} and 3 public Chinese datasets (Tnews, EPRSTMT and BUSTM) from FewCLUE \cite{fewclue}. We build all the few-shot learning datasets by sampling \textbf{20 samples} for each class to form training set. We also held out the same amount of samples for validation set but keep the whole test set unchanged. We train the model for 20 epochs and select the best model based on validation set. Table \ref{tab:exp-fewshot} shows that LaCon significantly outperforms the BERT-base fine-tuned with CE loss with a huge margin. Specifically, we observe 9.4\%, 7\%, and 6.2\% absolute improvement on BUSTM, QQP, and EPRSTMT. 

\begin{figure}[!htb]
\begin{center}
\subfloat[QQP]{%
\includegraphics[scale=0.16]{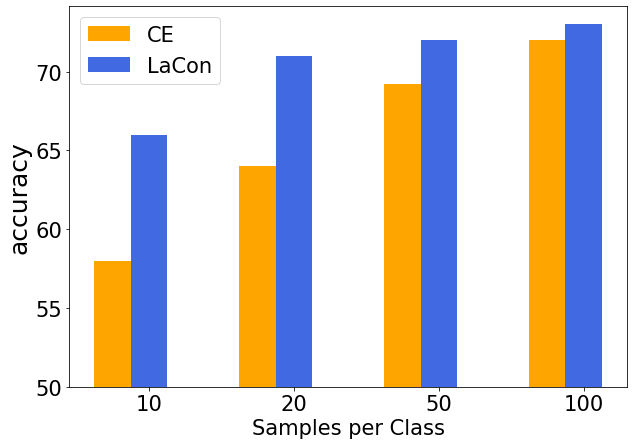}
}
\subfloat[RTE]{
\includegraphics[scale=0.16]{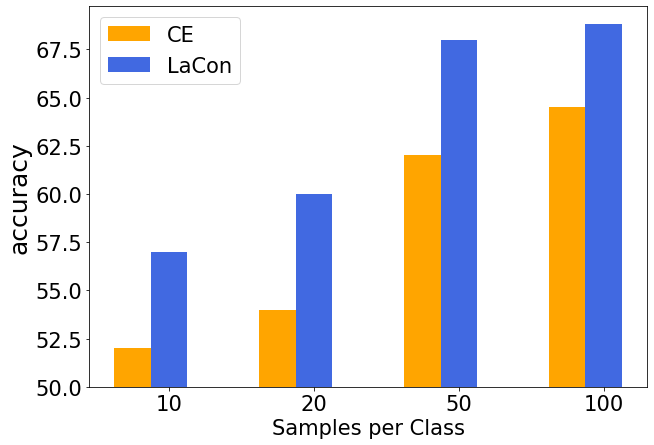}
}

\caption{Few-shot learning with different number of training samples.}
\label{fig:fewshot}
\end{center}
\vspace{-2mm}
\end{figure}

Additionally, we also conduct more strict experiments by changing the number of samples per class from \{10, 20, 50, 100\}. Figure \ref{fig:fewshot} demonstrates that the smaller the sample size per class is, the larger gain the model obtains. All above results indicate that the similarity-based CL losses in LaCon are able to hone in on the important dimensions of the multidimensional hidden representations hence lead to better and more stable few-shot learning results when fine-tuning PtLM.

\subsection{LaCon for Data Imbalance Setting}
\begin{figure}[!htb]
\begin{center}
\subfloat[QNLI]{%
\includegraphics[scale=0.26]{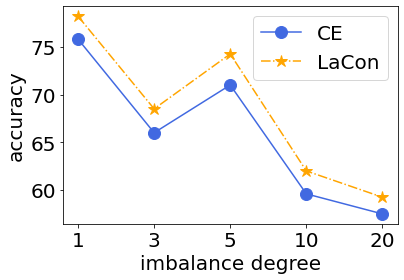}
}
\subfloat[CoLA]{
\includegraphics[scale=0.26]{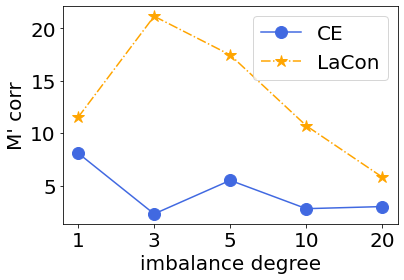}
}
\caption{LaCon with Different Imbalance Degree ($\rho$).}
\label{fig:imbalance}
\end{center}
\vspace{-2mm}
\end{figure}

The real-world datasets are usually imbalanced for different classes \cite{imbalancenips19, imblabdist20}, where several dominant classes contain most of the samples while the rest minority classes only hold a handful of samples. In this section, we conduct experiments to validate the capacity of LaCon under data imbalance setting. We follow the previous research \cite{imbalancenips19} to construct imbalanced classification training datasets with different imbalance degree ($\rho=|class_{max}|/|class_{min}|$, where $|class_{max}|$ / $|class_{min}|$ denotes number of samples in maximum / minimum class). For space limitation, we conduct experiments with vanilla LaCon on QNLI and CoLA. The minority class contains 32 samples and the majority class contains $32 \times \rho$ in our experiments. As shown in Figure \ref{fig:imbalance}, we vary the imbalance degree ($\rho$) from \{1, 3, 5, 10, 20\} and observe that LaCon outperforms BERT with CE consistently, demonstrating that LaCon also has advantage on the data imbalance setting.

\begin{table}[htp]
\centering
\small
\begin{tabular}{ccccc}
\hline
{} & \multicolumn{2}{c}{\textbf{QNLI}} & \multicolumn{2}{c}{\textbf{CoLA}}  \\
\hline 
\textbf{Methods} & \textbf{minor} & \textbf{major}  & \textbf{minor} & \textbf{major}   \cr
 \hline
CE & {43.4} & {71.8} & {2.3} & {81.3}  \cr
LaCon & {56.1} & {74.0} & {12.9} & {82.2} \cr
 \hline
\end{tabular}
\caption{F1 score for both majority and minority classes. Due to space limitation, we show results of $\rho=10$.}
\label{tab:detailed-imbalance}
\end{table}

We argue that LaCon may alleviate data imbalance issue on two aspects: 1) For the infrequent classes, treating labels as anchor or positive/negative may mitigate the data insufficient issue to some extent. 2) Label representations are shared across the whole dataset during training, which may transfer the knowledge from frequent classes to infrequent classes. To validate above conjecture, we present the performance on the test sets for majority and minority classes separately. Table \ref{tab:detailed-imbalance} shows that LaCon outperforms the baseline on both majority and minority classes and the gain on minority class is much larger.

\subsection{Visualization}

\begin{figure}[!htb]
\begin{center}
\subfloat[CE]{%
\includegraphics[scale=0.18]{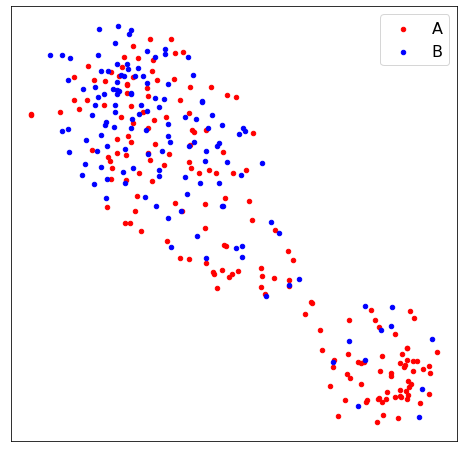}
}
\subfloat[LaCon]{
\includegraphics[scale=0.18]{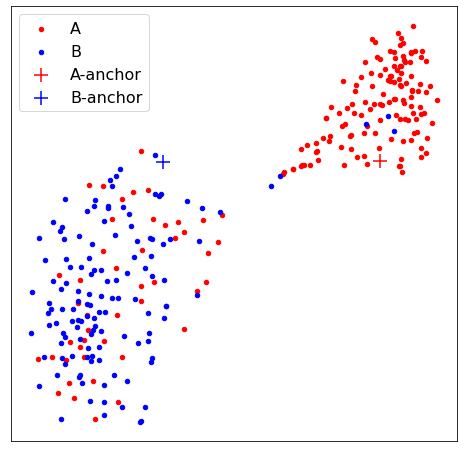}
}
\\
\subfloat[CE]{%
\includegraphics[scale=0.18]{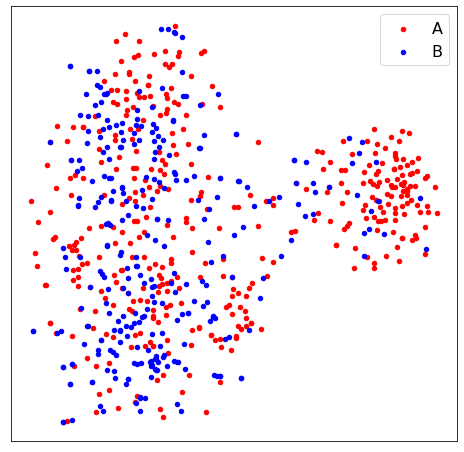}
}
\subfloat[LaCon]{
\includegraphics[scale=0.18]{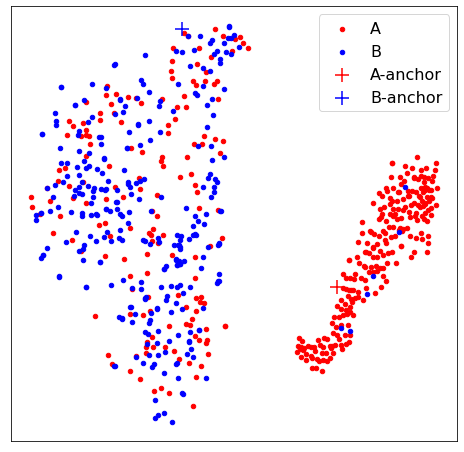}
}
\caption{Visualization of label and instance representations for MRPC (a\&b) and CoLA (c\&d) using T-SNE.}
\label{fig:visual}
\end{center}
\end{figure}

To demonstrate the effectiveness of LaCon on representation learning, we visualize the learned instance representations of LaCon and CE on the MRPC and CoLA dataset. In Figure \ref{fig:visual}, we use t-SNE \cite{van2008visualizing} to visualize both the high dimensional representations of the instances and labels on a 2D map. Different classes are depicted by different colors. As shown in Figure \ref{fig:visual} (a), the instances of class A and class B are sparsely located and overlapped in a large area, making it hard to find a hyper-plane to separate them. However, in Figure \ref{fig:visual} (b), the instances gather into two compact clusters and the instances stay close to the corresponding class. For CoLA, Figure \ref{fig:visual} (c) and (d) show the similar trends. It indicates that LaCon can learn more discriminative instance representations than CE. Besides, in Figure \ref{fig:visual} (b), the instances are near the corresponding label anchor, proving that LaCon can also learn a representative label embedding for each class.

\subsection{Hyper-parameter Tuning}
\label{app:hypertune}

\begin{figure*}[!htb]
\begin{center}
\subfloat[$\tau$]{%
\includegraphics[scale=0.28]{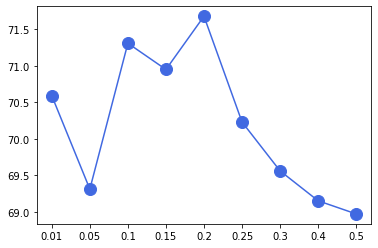}
}
\subfloat[$heads$]{
\includegraphics[scale=0.28]{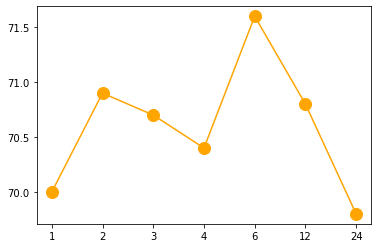}
}
\subfloat[$\lambda$]{
\includegraphics[scale=0.28]{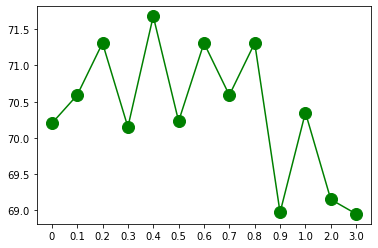}
}

\caption{Illustration of hyper-parameters tuning (RTE is taken for example and other datasets are similar).}
\label{fig:hyper}
\end{center}
\end{figure*}

In this section, we take the RTE dataset as an example for illustrating the hyper-parameter tuning process. The similar hyper-parameter tuning strategy is applied for other datasets. The tuning scripts will be released in our source code. Figure \ref{fig:hyper} shows the influence of different hyper-parameters. 

For each experiment, we conduct a grid-based hyper-parameter sweep for $\tau$ between 0.05 and 0.5 with step 0.05, $\lambda$ between 0.1 and 1.0 with step 0.1, and select the best hyper-parameter for the given dataset. The $\tau$ is the most influential hyper-parameter that needs to be tuned carefully with minimum step 0.05. Larger $\tau$ results in lower accuracy in LaCon and the recommended value is around 0.1 and 0.2. Figure \ref{fig:hyper} (b) illustrates that the optimal number of heads in Equation \ref{eq:mh-loss} is 6 and both the most and fewest heads result in low accuracy while heads with middle sizes get relatively better accuracy scores. Small number of head shows little diversity in feature clipping while larger one results in very short vectors with poor representation capacity. The label embedding regularizer weight $\lambda$ in Figure \ref{fig:hyper}(c) can be set in a wide range, where either without $\mathcal{L}_{LER}$ or large $\lambda$ will result in poor performance.

\subsection{The Impact of Class Number}
\label{app:numlabels}

\begin{figure}[!htb]
\begin{center}
\includegraphics[scale=0.28]{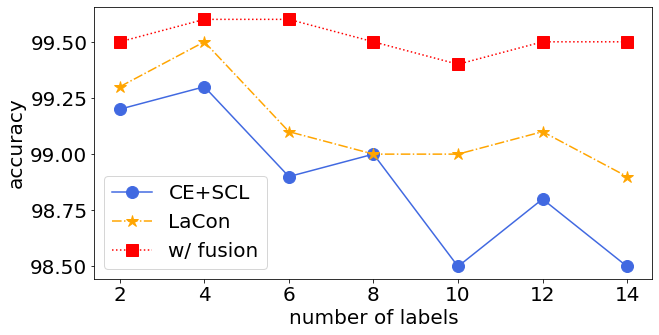}
\caption{The impact of class number on LaCon. Experiments were conducted on DBPedia.}
\label{fig:label-count}
\end{center}
\end{figure}

As the number of classes influence the difficulty of classification task directly, in this section, we discuss the impact of class number on our proposed model LaCon. We pick the DBPedia dataset for experiment. The original DBPedia dataset includes 14 labels. We gradually increase the label number from 2 to 14 and randomly select 1000 samples for each label in our experiment as training set. Meanwhile, we keep the whole samples for the chosen labels in the evaluation set unchanged. Figure \ref{fig:label-count} demonstrates that with the increase of the labels, the performance of all models degrades as the task becomes more difficult. However, LaCon-fusion outperforms CE+SCL consistently on different number of labels, which shows the advantage of leveraging labels as anchors or positive/negative samples during contrastive learning.

\section{Related Work}

\subsection{Contrastive Learning}

Contrastive Learning has become a rising domain and achieved significant success in various CV, speech and NLP tasks \cite{Moco2019,chen2020simple, cert21, speechtext-acl21, contrast-audio-general-repr/icassp/SaeedGZ21, sclce21, simcse21, ConSERT21}. There are two kinds of CL approaches, which are self-supervised CL and supervised CL. The self-supervised CL contrasts a single positive for each anchor (i.e., an augmented version of the same image) against a set of negatives consisting of the entire remainder of the batch. However, due to the intrinsic discrete nature of natural language, data augmentations are less effective than that in CV. Recently, researchers \cite{scl20, sclce21} propose supervised CL, which contrasts the set of all samples from the same class as positives against the negatives from the remainder of the batch. \citet{lcl21} propose label-aware SCL method via assigning weights to instances of different labels, which treats the negative samples differently.

LaCon belongs to the scope of supervised CL. Different from \cite{scl20, sclce21}, LaCon can take the labels as anchors or mine negative/positive from labels, which does not need to construct positive pairs from the data augmentation. Meanwhile, \citet{sclce21} combine CL and CE losses at the same time, but LaCon is purely equipped with three CL objectives, including the instance-centered contrastive loss, the label-centered contrastive loss and the label embedding regularizer.

\subsection{Label Representation Learning}
Label representation learning aims to learn the embeddings of labels in classification tasks and has been proven to be effective in various CV \cite{Frome2013,Akata2016} and NLP tasks \cite{Tang2015,Pappas2019,nam2016all,zhang2017multi,wang2018joint,lsan-xiao2019label,Miyazaki2020}. In this work, we compare with two representative label embedding based models, which are LEAM \cite{wang2018joint} and LSAN \cite{lsan-xiao2019label}. Both learn label embeddings and sentence representations in a joint space based on attention mechanism and fuse them to improve the classification. Differently, LaCon learns the label and instance representations jointly via purely supervised contrastive learning. Besides, our experiments also verify that after obtaining the discriminative label and instance representations, even simple fusion block can facilitate the language understanding tasks.

\section{Conclusions}
In this paper, we proposed a novel supervised contrastive learning approach for language understanding. To utilize the class labels sufficiently, we devise three novel contrastive objectives, including a multi-head instance-centered contrastive loss, a label-centered contrastive loss, and a label embedding regularizer. Extensive experiments were conducted on eight public datasets from GLUE and CLUE benchmarks, showing the competitiveness of LaCon against various strong baselines. Besides, we also demonstrate the strong capacity of LaCon on more challenging few-shot and data imbalance settings, which leads up to 9.4\% improvement on the FewGLUE and FewCLUE benchmarks. LaCon does not require any complicated network architecture or any extra data augmentation, and can be easily plugged into mainstream pre-trained language models. In the future, we will explore more advanced representation fusion approaches to enhance the capability of LaCon and plan to extend LaCon to the computer vision and speech fields.


\bibliography{anthology}

\appendix
\section{Appendix}

\subsection{Accumulative Ablation Study}
\label{app:accum_ablation}

In this section, we supplement more ablation results by adding each proposed CL loss cumulatively. We conduct experiments on MRPC, RTE, and CoLA datasets and keep the setting consistent with Section \ref{sec:ablation}. Table \ref{tab:cumu-ablation} demonstrates that the contribution of each component in more details. 
\begin{table}[htp]
\centering
\small
\begin{tabular}{ccccc}
\hline \textbf{Methods} & \textbf{MRPC} & \textbf{RTE} & \textbf{CoLA} \\ 
\hline
$\mathcal{L}_{ICL}$ & {86.2$\pm1.5  $} & {67.8$\pm1.1 $} & {61.2$\pm0.9 $}  \cr
$\mathcal{L}_{ICL}^{'}$ & {86.9$\pm1.1  $} & {68.2$\pm0.7 $} & {61.2$\pm1.5 $}  \cr
$\mathcal{L}_{LCL}$ & {87.0$\pm1.7  $} & {68.3$\pm1.0 $} & {61.1$\pm1.3 $}  \cr
$\mathcal{L}_{ICL}^{'}+\mathcal{L}_{LCL}$ & {87.3$\pm 1.1 $} & {70.2$\pm 1.3$} & {62.2$\pm 1.6$}  \cr
$\mathcal{L}_{ICL}^{'}+\mathcal{L}_{LER}$ & {87.1$\pm 0.6 $} & {70.5$\pm 0.8$} & {61.2$\pm 1.1 $} \cr
\hline
\textbf{LaCon-vanilla} &  \textbf{87.5$\pm \textbf{0.8} $} & \textbf{71.4$\pm \textbf{0.7} $} & \textbf{62.4$\pm \textbf{1.1} $}  \cr
\hline
\end{tabular}
\caption{Ablation study via adding losses cumulatively.}
\label{tab:cumu-ablation}
\end{table}

\subsection{Experimental Results of More Datasets}
\label{app:glue-complete}

We supplement more experimental results on the remaining datasets of GLUE and CLUE benchmarks. We follow the same experimental setup with Section \ref{sec:exp-setup}. Please note that SST-B is a regression task that is beyond the capacity of the proposed LaCon. The official CLUE benchmark has replaced CMNLI with OCNLI dataset, and the CSL dataset is a keyword recognition task, which is not suitable for our proposed model. Thus, we omit the experiments on above three datasets and report the performance on the remaining language understanding tasks including SST-2, MNLI, AFQMC, OCNLI and IFLYTEK. 

Table \ref{tab:glue-addon} shows that, LaCon-vanilla consistently outperforms BERT fine-tuned with CE, and LaCon-fusion still beats the baselines among all datasets, which further demonstrates the superiority of our proposed method. 

\begin{table*}[htp]
\centering
\small
\begin{tabular}{cccccc}
\hline
\textbf{Methods} & {\textbf{SST-2}}  & \textbf{MNLI} & \textbf{AFQMC} & \textbf{OCNLI} & \textbf{IFLYTEK} \cr
 \hline
{Baseline*} & {91.4$\pm0.3$} & {73.1$\pm0.3$} & {74.5$\pm0.3$}  & {68.3$\pm0.7$} & {61.5$\pm0.5$}  \cr
{CE} & {91.2$\pm0.1$} & {72.5$\pm0.3$} & {70.9$\pm0.5$} & {66.8$\pm0.4$} & {60.6$\pm0.2$} \cr
{LaCon-vanilla} & {91.4$\pm0.3$} & {73.3$\pm0.4$} & {72.5$\pm0.3$} & {67.4$\pm0.3$} & {60.9$\pm0.2$} \cr
{\textbf{LaCon-fusion}} & {\textbf{92.5$\pm\textbf{0.2}$}} & \textbf{{73.9$\pm\textbf{0.3}$}} & \textbf{{74.8$\pm\textbf{0.5}$}} & \textbf{{69.1$\pm\textbf{0.6}$}} & \textbf{{63.5$\pm\textbf{0.7}$}} \cr
\hline
\end{tabular}
\caption{Performance on the remaining datasets of GLUE and CLUE. Baseline* means the best performance of our baselines. The evaluation metrics are the same as the official GLUE \cite{2019-GLUE} and CLUE \cite{2020-Clue} benchmarks (all with significance value $p < 0.05$).}
\label{tab:glue-addon}
\end{table*}

\subsection{Theoretical Analysis}
\label{app:theory}
In this section, we conduct the theoretical analysis to prove the rationality and necessity of our proposed ICL and LCL losses. We also explain why these two losses are complementary. Finally, we analyze the computational efficiency of ICL and LCL compared to InfoNCE \cite{infoNCE18} and SCL \cite{scl20}. 

The recent researches \cite{bertflow2020, repr-degen-2019iclr} reveal that the anisotropy problem of pre-trained language models, which shows that the learnt embeddings occupy a narrow cone in the dense vector space, harming the uniformity of the models and limiting the representation capacity. The singular values of the contextual embeddings decay drastically with most of them nearly zeros \cite{ctl-theory20}. CL is proposed to eliminate the long-tail distribution problem of singular values, aiming to enhance the representation capacity \cite{simcse21, ConSERT21, sclce21}.
From the spectrum perspective \cite{spectrumctr20iclr, ctl-theory20} that analyzes the distribution and uniformity of the learned embedding space, CL flattens singular values of the embeddings thus improves the capacity of language models.
\begin{equation}
\label{eq:theorem-inst}
\begin{split}
\mathcal{L}_{ICL} = 
-\frac{1}{\tau} E_{(x,y) \sim A(y)}(H_xL_y ) 
\\
+E_{x \sim A(y)} [log E_{y^{-} \notin A(y)}(e^{H_xL_{y^-}/\tau}) ]
\\
\end{split}
\end{equation}

\begin{equation}
\label{eq:theorem-inst-jensen}
\begin{split}
E_{x \sim A(y)} [log E_{y^{-} \notin A(y)}(e^{H_xL_{y^-}/\tau}) ]
\\
= \frac{1}{N}\sum_{i=1}^{i=N}log(\frac{1}{C-1}\sum_{ j\neq i}^{1\leq j \leq C}e^{H_{x_i} L_{y_j}/\tau} )
\\
\geq \frac{1}{N(C-1)\tau} \sum_{i=1}^{i=N}\sum_{j=1, j\neq i}^{j=C}H_xL_{y^-}
\\
\end{split}
\end{equation}

Therefore, we can form an asymptotic equivalent objective of the $\mathcal{L}_{ICL}$ (Equation \ref{eq:inst-loss}) as Equation \ref{eq:theorem-inst}.
$(x,y) \sim A(y)$ denotes instances (i.e. $x$) with corresponding label (i.e. $y$) and $y^-$ denotes the label that is different from $y$. The first item keeps instances and corresponding labels similar and the second item pushes the mismatched instances and labels apart. We can further derive Equation \ref{eq:theorem-inst-jensen} using Jensen's inequality because $e(.)$ is convex. Therefore, minimizing the $\mathcal{L}_{ICL}$ equals to minimization of summation of all elements in $HL^T \in R^{N \times C}$. Because both $H$ and $L$ are normalized, $tr(H^TL)$ is a constant due to all diagonal elements are ones. $sum(HL^T)$ is an upper bound of the largest singular value \cite{MERIKOSKI1984177} and minimization of the $sum(HL^T)$ will flatten the singular values distribution of $HL^T$. As the $HL^T$ is a non-squared matrix, we need to optimize both the left and right singular values using $HL^T$ and $LH^T$ in order to effectively eliminate the anisotropy and promote the uniformity of pre-trained language models in classification tasks to enhance the model capacity. Thus, we also need to optimize the label-centered contrastive loss $\mathcal{L}_{LCL}$ at the same time. From above analysis, we can see that $\mathcal{L}_{ICL}$ and $\mathcal{L}_{LCL}$ are complementary to each other. Similarly, we can derive that minimizing $\mathcal{L}_{LCL}$ results in the minimization of $sum(LH^T) \in R^{C \times N}$.

Although both ICL and LCL calculate the $N\times C$ similarity scores for a mini-batch, they are different. The ICL is the average of the instance-level per sample loss while the LCL is the per label loss. The ICL intends to align each instance to corresponding label correctly. The LCL makes the instances of different labels far away from each other and instances of the same label more compact. They consider different aspects of instance and label representation through operating the $N \times C$ similarity scores differently according to Equation \ref{eq:inst-loss} and \ref{eq:label-loss}.

Compared to InfoNCE, ICL improves the computational efficiency from $O(N^2)$ to $O(NC)$ because we only need to contrast the instance representation with corresponding label representations for per example loss, which is extremely useful for language understanding tasks \cite{2019-GLUE, 2020-Clue} that commonly consist of 2 or 3 labels. Similarly, LCL is also more computationally efficient as it only contrasts one label representation to several instances rather than computes all pairs of instances belonging to a given label in the mini-batch. Thus it improves the complexity from $O(N^2)$ to $O(CN)$ compared with SCL too.

\end{document}